\documentclass{article}

\usepackage{graphics}
\usepackage{amsmath}
\usepackage{amssymb}
\usepackage{mathtools}
\usepackage{amsthm}
\usepackage{multirow}
\newtheorem{theorem}{Theorem}
\newtheorem{definition}{Definition}
\usepackage{subcaption}
\usepackage[numbers]{natbib}
    \usepackage[preprint]{neurips_2025}

\usepackage[utf8]{inputenc} % allow utf-8 input
\usepackage[T1]{fontenc}    % use 8-bit T1 fonts
\usepackage{hyperref}       % hyperlinks
\usepackage{url}            % simple URL typesetting
\usepackage{booktabs}       % professional-quality tables
\usepackage{amsfonts}       % blackboard math symbols
\usepackage{nicefrac}       % compact symbols for 1/2, etc.
\usepackage{microtype}      % microtypography
\usepackage{xcolor}         % colors

\title{Rethinking the Understanding Ability across LLMs through Mutual Information}

\author{%
  \begin{tabular}{c c}
    \begin{minipage}{0.45\textwidth}
      \centering
      \textmd{Shaojie Wang} \\
      \textmd{Department of Industrial and Systems Engineering} \\
      \textmd{University of Houston} \\
      \texttt{swang86@uh.edu}
    \end{minipage}
    &
    \begin{minipage}{0.45\textwidth}
      \centering
      \textmd{Sirui Ding} \\
      \textmd{Bakar Computational Health Sciences Institute} \\
      \textmd{UCSF} \\
      \texttt{sirui.ding@ucsf.edu}
    \end{minipage}
  \end{tabular}
  \\[20pt]
  \begin{tabular}{c}
    \textmd{Na Zou} \\
    Department of Industrial and Systems Engineering \\
    University of Houston \\
    \texttt{nzou2@central.uh.edu}
  \end{tabular}
}

\begin{document}

\maketitle

\begin{abstract}
Recent advances in large language models (LLMs) have revolutionized natural language processing, yet evaluating their intrinsic linguistic understanding remains challenging. Moving beyond specialized evaluation tasks, we propose an information-theoretic framework grounded in mutual information (MI) to achieve this. We formalize the understanding as MI between an input sentence and its latent representation (sentence-level MI), measuring how effectively input information is preserved in latent representation. Given that LLMs learn embeddings for individual tokens, we decompose sentence-level MI into token-level MI between tokens and sentence embeddings, establishing theoretical bounds connecting these measures. Based on this foundation, we theoretically derive a computable lower bound for token-level MI using Fano's inequality, which directly relates to token-level recoverability-the ability to predict original tokens from sentence embedding. We implement this recoverability task to comparatively measure MI across different LLMs, revealing that encoder-only models consistently maintain higher information fidelity than their decoder-only counterparts, with the latter exhibiting a distinctive late-layer "forgetting" pattern where mutual information is first enhanced and then discarded. Moreover, fine‑tuning to maximize token-level recoverability consistently improves understanding ability of LLMs on tasks without task‑specific supervision, demonstrating that mutual information can serve as a foundation for understanding and improving language model capabilities.
\end{abstract}

\section{Introduction}
Recent advances in large language models (LLMs) have catalyzed a paradigm shift in generative AI, allowing it to move beyond simple pattern matching to exhibit behaviors resembling reasoning and comprehension \cite{llm_reasoning_survey,llm_understanding}. As these models rapidly diversify in architecture (e.g., dense networks vs. mixture-of-experts\cite{moe}) and scale (from 1B to over 100B parameters\cite{scale_llm}), comparing these diverse models becomes increasingly challenging, as their fundamental differences in design and training may influence performance in ways unrelated to their actual linguistic capabilities. Given this diversity, a critical question emerges: How can we evaluate and compare their intrinsic linguistic understanding capabilities in an architecture-agnostic manner?"

The current evaluation paradigm primarily relies on performance metrics from a series of natural language understanding tasks\cite{survey_eval_llm,exp_encoder_decoder,nlu_multiple_factors}. These evaluations attempt to quantify a model's comprehension abilities by measuring its performance on carefully curated tasks, including Named Entity Recognition \cite{ner}, Word Sense Disambiguation and Words-in-Context tasks \cite{encoder_understanding_better}, and text classification \cite{exp_encoder_decoder}, etc. Although these approaches have yielded valuable empirical insights, recent studies have pointed out several limitations of this paradigm, such as knowledge conflicts\cite{constraint_nlu_task}, vulnerability\cite{llm_vulnerability},  misrepresentation\cite{Limitation_benchmark,benchmark_poor} and effects related to model scale\cite{llm2vec_compare}. Moreover, task-centric evaluations are highly dependent on specific datasets and task formulations, making it challenging to determine whether performance differences stem from genuine understanding capabilities or merely from task-specific advantages.

% \textbf{1)} They exhibit strong dependence on task-specific datasets and evaluation protocols, limiting generalization across diverse linguistic environments \cite{constraint_nlu_task}. Models often perform inconsistently across different benchmarks, making it challenging to draw reliable conclusions about their general understanding capabilities\cite{llm_vulnerability,Limitation_benchmark}. \textbf{2)}. These evaluations are susceptible to confounding factors—including pre-training objectives, tokenization strategies, prompt order, and model scale\cite{benchmark_poor}—that can substantially influence performance metrics without necessarily reflecting differences in the models' intrinsic language comprehension abilities \cite{llm2vec_compare}. As a result, it remains challenging to determine whether performance differences stem from genuine understanding capabilities or others, or like merely from task-specific advantages, etc. 

Given these limitations, there is a clear need for an evaluation framework that can assess language understanding independently of task-specific performance or architectural details. An ideal framework should provide a principled measure that captures the essence of understanding: the ability to extract and preserve information from language inputs. 

% Inspired by cognitive neuroscience, where mutual information (MI) quantifies the integration of information in neuronal populations \cite{information_population_2,information_population}, we argue that the information theory principles applied to biological neural networks can be analogously applied to artificial ones, providing a more principled framework for evaluation.

\begin{figure} \centering\includegraphics[width=\linewidth,height=0.7\textheight,keepaspectratio]{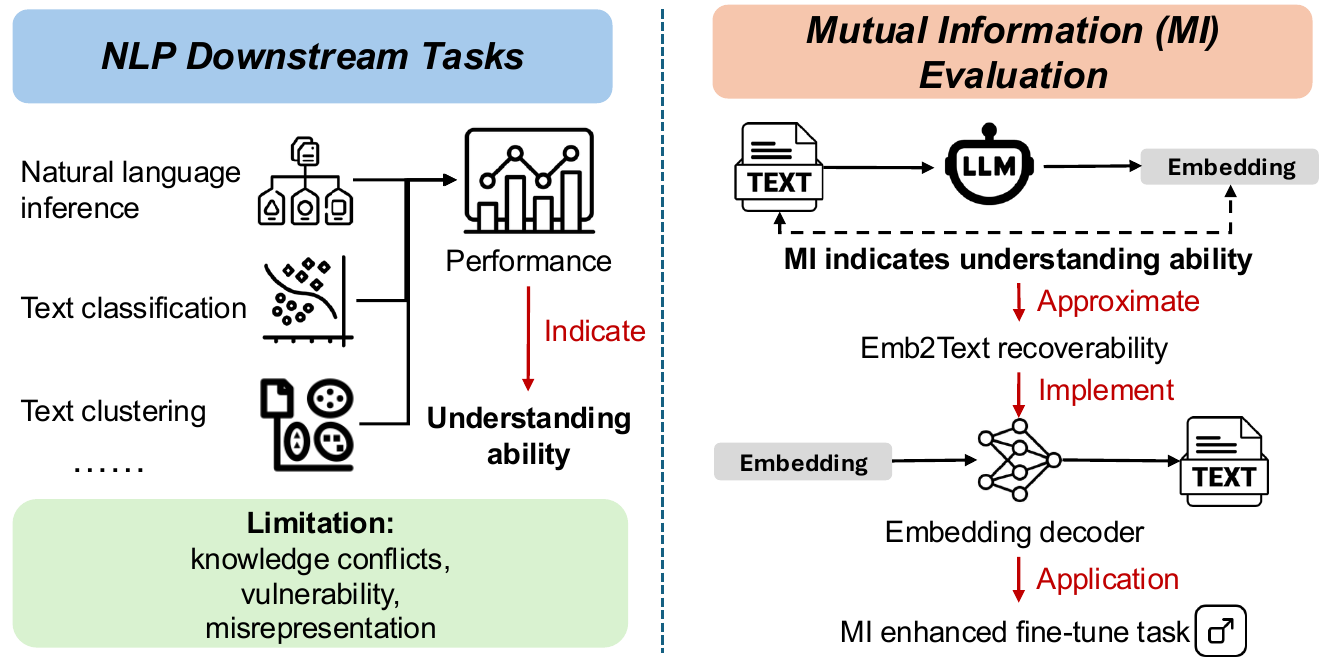}
  \caption{Two paradigms for evaluating language understanding in LLMs. \textbf{Left}: Traditional task-specific evaluation relies on performance from diverse NLP tasks (natural language inference, text classification, text clustering, etc.), which can meet several limitations. \textbf{Right}: Our proposed framework provides an architecture-agnostic approach that measuring information preservation in model representations through mutual information, where token-level recoverability is served as a practical implementation. Then we enhance MI based on our framework by increasing token-level recoverability, which can increase understanding ability of LLMs.}
  \label{structure}
 \vskip -0.1in
\end{figure}

Specifically, we define the language understanding capacity of a model through the mutual information between input sentences and their latent representations (sentence-level MI). By measuring how effectively input information is preserved in latent representations, similar to how neuroscientists analyze stimulus encoding in neural circuits \cite{mutual_information_neuronal_circuits}, we can gain deeper insight into how different LLMs retain linguistic information. Unlike task-specific evaluations, this approach offers a task-agnostic and model-agnostic insight that can be applied to LLMs with different architectures and training objectives. However, directly computing MI in LLMs presents challenges. The high dimensionality of the neural representations makes density estimation difficult \cite{difficulty_mutual_information}.

To overcome the above challenges, we develop a computable and decomposable framework to rigorously approximate and estimate the understanding ability of LLM, i.e., sentence-level MI. Considering that LLMs process input sentences and generate embeddings for individual tokens, we decompose sentence-level MI into more tractable token-level components, specifically examining the mutual information between individual tokens and sentence embeddings (token-level MI), then establishing theoretical bounds between sentence-level MI and token-level MI. Based on these theoretical foundations, we derive a computationally lower bound for token-level MI with theoretic guarantees grounded in Fano's inequality\cite{fano_inquality}. This bound corresponds to token-level recoverability—the ability to accurately predict original tokens from sentence embeddings. This theoretically derived lower bound allows us to systematically measure and compare understanding ability across various LLMs. The detailed structure of our proposed framework is illustrated in Figure \ref{structure}, which contrasts our information-theoretic approach with traditional task-specific evaluation paradigms to assess language understanding in LLMs.

Our framework positions token recoverability as both an estimation method for token-level MI and an indicator of information preservation. When a model effectively preserves token information in its embeddings, it retains more fine-grained linguistic details necessary for deeper language understanding. By analyzing token recoverability across different layers of LLMs with different architectures (encoder-only vs. decoder-only), we reveal distinct patterns in how information flows, with the latter exhibiting a distinctive late-layer "forgetting". Furthermore, we empirically validate our theoretical framework by enhancing token-level recoverability which is also understanding ability through fine-tuning, which can consistently improve performance across multiple downstream tasks without task-specific adaptation.

Our contributions are as follows: \textbf{1)}.We present a task-agnostic and model-agnostic quantification framework to analyze LLM comprehension, establishing sentence-level MI as a fundamental metric for evaluating semantic preservation and understanding in latent representations. \textbf{2)}. Through a theoretical investigation, we can approximate sentence-level MI via a lower bound, i.e., the predefined token-level recoverability, which can systematically measure layer-wise understanding and thus the information flow of LLMs. \textbf{3)}.Through extensive experimentation, we empirically validate the effectiveness of the proposed framework on quantifying and measuring the understanding of LLMs, as well as to further fine-tune the models for consistent performance improvement on language understanding tasks.

\section{Related Work}
\subsection{Task-specific evaluation of LLMs}
Multiple works compare the understanding capacity between different LLMs by leveraging specialized tasks. For example, word-in-context and word-sense disambiguation evaluations reveal that decoder-only LLMs perform worse in capturing word meanings than their encoder-only counterparts\cite{encoder_understanding_better}. In addition, extensive experiments have been conducted to investigate performance differences across various LLMs on financial text analysis\cite{exp_encoder_decoder} and text classification tasks\cite{exp_encoder_decoder_2}, while further studies examine different architectures in code search\cite{codesearch_compare} or embedding-based approaches (e.g., LLM2Vec\cite{llm2vec_compare}). Through an extended ScandEval benchmark, people found that encoder models significantly outperformed decoder models on multilingual nature language understanding tasks despite having orders of magnitude fewer parameters\cite{multi_lingual}. By providing snippets of text and querying the model's interpretation, they question whether LLMs truly 'understand' language\cite{llm_lack_understanding}. A benchmark called SUC is created to evaluate the level of understanding that large language models (LLMs) have of structured table data\cite{understanding_table}.
% All these task-centric studies rely heavily on specific experimental designs and datasets, which highlights the need for a more unified framework capable of evaluating how effectively models preserve and encode linguistic information, independent of any single task or domain.
% BioMistral-NLU\cite{biomistral} curated an instruction-tuning dataset to address the performance gap between instruction-tuned LLMs and specialized models on medical natural language understanding tasks. By combining machine translation encoders with LLM backbones through self-distillation, LLMs' natural language understanding capabilities are extended to low-resource languages\cite{cross_lingual}.

\subsection{Mutual information}
MI has found applications in a variety of scientific fields, including astrophysics\cite{astrophysics}, biophysics\cite{biophysics} and dynamical complex systems\cite{complex_system}, and it has increasingly been used as a theoretical tool to investigate how well neural networks encode and preserve information from input data throughout their latent representations. By studying mutual information throughout stochastic deep learning, the relationship between compression and generalization remains elusive\cite{mi_dnn}. The visualization of MI in Deep learning networks can reflect the roles of Deep learning units in classification performance of the networks\cite{mi_vis}. In knowledge distillation scenarios, by maximizing the MI of the teacher model, the teacher model can capture richer contextual dependencies and pass more information to the student model\cite{mi_enhance}. By leveraging a mutual information-based objective to select words from a fixed vocabulary set, people demonstrated an effective approach for explaining sparse autoencoder features in LLMs\cite{llm_mutual_interpret}. MI-DPO\cite{mutual_information_dpo} uses mutual information principles and learnable priors to show DPO variants are actually special cases of a single.

\section{An Information-Theoretic Framework}
In this section, we introduce our information-theoretic framework. We first establish the relationship between token-level recoverability and token-level mutual information, then connect token-level mutual information to sentence-level mutual information.

\subsection{From Tokens to Sentences: Theoretical Bounds on Mutual Information}
In this section, we show how mutual information between individual tokens and the embedding of the sentence influence the sentence-level mutual information.

We now derive a lower bound on the mutual information $I(S;E)$ between the sentence $S$ and its embedding $E$ in terms of the average token-level mutual information.
\begin{theorem}[Compositional Information Lower Bound]
\label{theorem:lower bound}
For a sentence \(S = (t_1, t_2, \ldots, t_n)\) and its embedding \(E\), the following relationship holds between the sentence-level mutual information and the average token-level mutual information:
\begin{equation}
I(S; E) \geq n \cdot \frac{1}{n}\sum_{i=1}^{n} I(t_i; E) + \Delta
\end{equation}
where \(\Delta_=H(t_1, t_2, \ldots, t_n) - \sum_{i=1}^{n} H(t_i)\), represents the difference between the joint entropy of all tokens in sentence $S$ and the sum of their individual entropies.

\end{theorem}
This theorem establishes that the mutual information between a sentence and its embedding is lower-bounded by expressions involving the average token-level mutual information. Detailed proofs are provided in Appendix \ref{MI_sentence_token}.

\subsection{Quantifying Mutual Information via Recoverability}
In this section, we formalize the relationship between token-level recoverability and token-level mutual information.

\begin{definition}[Token-level Recoverability]
\label{def:token_recoverability}
For a sentence $S = (t_1, t_2, ..., t_n)$ and its embedding $E$, we define token-level recoverability as the probability of correctly recovering individual tokens in the sentence from its embedding using a mapping function:
\begin{equation}
P_{\mathrm{rec}}(S, E) := \max_{f} \frac{1}{n}\sum_{i=1}^{n}\mathbb{P}\Bigl[f(E)_i = t_i\Bigr]
\end{equation}
where $f$ represents any measurable function that attempts to reconstruct the original tokens from the embedding, $\text{i.e. } f : \mathbb{R}^d \to \mathcal{S} \text{ with } d = \dim(E)$. $E$ is the embedding before the last layer in LLM, since the final layer directly uses this embedding to perform tasks. $f(E)_i$ denotes the prediction of $f$ for the $i$-th token.
\end{definition}

\begin{theorem}[Token-level Recoverability and Mutual Information]
\label{thm:token_mi_bound}
For a sentence $S = (t_1, t_2, ..., t_n)$ with tokens drawn from vocabulary $\mathcal{V}$, assuming tokens are uniformly distributed over $\mathcal{V}$ (i.e., $H(t_i) = \log |\mathcal{V}|$), for large vocabularies (i.e., $|\mathcal{V}| \gg 1$), the mutual information between each token and the sentence embedding satisfies:
\begin{equation}
\frac{1}{n}\sum_{i=1}^{n} I(t_i; E) \ge P_{\mathrm{rec}}(S, E)\log(|\mathcal{V}|-1) - H_b\bigl(P_{\mathrm{rec}}(S, E)\bigr)
\end{equation}
where $H_b$ is the binary entropy function defined as $H_b(p) = -p\log p - (1-p)\log(1-p)$.
\end{theorem}
This theorem establishes that higher token-level recover accuracy corresponds to higher average mutual information between individual tokens and the sentence embedding. Intuitively, when a model can accurately predict tokens from the sentence embedding, it demonstrates that the representation preserves substantial information about the original tokens. The detailed proof is provided in Appendix \ref{MI_rec}.

\section{Evaluation Task Design}
\label{task_setting}
This section details our experimental methodology for measuring and improving mutual information in LLM. We present two approaches: (1) a token-level recoverability task that serves as a practical implementation for measuring mutual information, and (2) a fine-tuning approach designed to enhance the understanding ability of LLMs.

\subsection{Token-level recoverability task}

\textbf{Task formulation}
We formulate token-level recoverability as a supervised classification problem, where we train an embedding decoder $f$ to reconstruct the original sentence $S$ from its embedding $E$. For each token $t_n$ in sentence  $S = (t_1, t_2, ..., t_n)$, we define its one-hot representation as: $o^{\text{one-hot}}(t) = 
\begin{cases}
1, & \text{if } t = t_n \\
0, & \text{otherwise}
\end{cases}
, \quad t \in \mathcal{V}$. The ground truth label matrix for the sentence $S$ is constructed as: $O = [o^{\text{one-hot}}(t_1); o^{\text{one-hot}}(t_2); ...; o^{\text{one-hot}}(t_n)] \in \mathbb{R}^{n \times |\mathcal{V}|}$. After the training process, we use well-trained $f$ to recover texts in different domains to verify mutual information.

According to previous findings\cite{llm_linear}, there is a high linear relationship between embeddings across different layers in LLM, we employ a linear transformation as our embedding decoder $f: \mathbb{R}^d \rightarrow \mathbb{R}^{|\mathcal{V}|}$ to map the embedding space to token probabilities: $f(E) = W \cdot E + b$, where $W \in \mathbb{R}^{|\mathcal{V}| \times d}$ and $b \in \mathbb{R}^{|\mathcal{V}|}$ are parameters of $f$. While more complex non-linear architectures could also be implemented, our linear approach provides an effective and computationally efficient solution. We train the decoder $f$ using Binary Cross-Entropy with Logits Loss:
\begin{equation}
\label{bce_loss}
\mathcal{L}(f(E), O) =  -\sum_{i=1}^{n}\sum_{t \in \mathcal{V}}[O_{i,t} \cdot \log(\sigma(f(E)_{i,t}))\\ +
(1 - O_{i,t}) \cdot \log(1 - \sigma(f(E)_{i,t}))]
\end{equation}

where $\sigma$ is the sigmoid activation function: $\sigma(f(E)) = \frac{1}{1+e^{-f(E)}}$. After $f$ is well-trained, we test the recoverability of different LLMs on texts from different domains, by generating embeddings first and then recover using well-trained $f$.

\textbf{Models and Datasets}
For this task, we train embedding decoders for each large language model (LLM) to evaluate its performance. We examine a diverse set of models spanning encoder-only and decoder-only architectures: 1).Encoder-only models: RoBERTa-large\cite{roberta}, DistilBERT\cite{distilbert}; 2).Decoder-only models: OPT\cite{opt}, Mistral\cite{mistral_7b}, Falcon\cite{falcon}, GPT-NEO-2.7b\cite{gpt_neo},LLaMA2-7b\cite{llama2}.

We use dbpedia\cite{yelp_amazon} dataset to train the embedding decoders. To comprehensively evaluate the recoverability performance across different domains and text types, we utilize multiple text datasets for testing: 1).News domain: AG\_News\cite{ag_news}; 2).Review domain: IMDB\cite{imdb}, Yelp\cite{yelp_amazon}, Amazon\cite{yelp_amazon}; 3). Question-answer domain: QNLI\cite{glue}; 4).Machine-generated text: ChatGPT-generated texts. During the test phase, we randomly select 200 texts from each dataset to evaluate the recovery performance of different LLMs.

\textbf{Evaluation Metrics}:
To comprehensively evaluate sentence recoverability performance, we employ multiple metrics: 1).Cosine similarity: Measures the similarity between the original and recovered sentence embeddings; 2).Token F1 score: Evaluates the precision and recall of sentence recovery; 3).BLEU-\{1,2,4\}: Assesses the n-gram overlap between original and recovered texts; 4).ROUGE-\{1,L\}: Measures the overlap of unigrams and longest common subsequences.

\subsection{Understanding enhanced fine-tuning}

\textbf{Task formulation}
We formulate the understanding enhanced task following the procedure below:
1). For a given LLM, we freeze the parameters of the final layer to maintain the model's existing output mapping.
2). We then optimize the parameters of all the preceding layers using the same recoverability task described in our former section. Formally, our optimization can be expressed as:
\begin{equation}
\theta_{1:L-1}^* = \arg\min_{\theta_{1:L-1}} \mathcal{L}_{\text{recov}}(f(E_{\theta_{1:L-1}}), O)
\end{equation}
where $\theta_{1:L-1}$ represents the parameters of all layers except the final layer, and $E_{\theta_{1:L-1}}$ is the embedding produced by these layers.

\textbf{Models and Datasets}
For the mutual information enhancement task, we focus on three LLMs with varying architectures and sizes: LLaMA2-7b\cite{llama2}; OPT-2.7b\cite{opt}; Mistral-7b\cite{mistral_7b}. We conduct our fine-tuning on the WikiText corpus\cite{wikitext}, which provides a diverse collection of high-quality articles suitable for enhancing mutual information across different contexts and domains. To verify effectiveness, four different tasks are evaluated: 1).Classification: AG-News\cite{ag_news}; 2).Semantic Textual Similarity: STS-B dataset from GLUE benchmark\cite{glue}; 3).Retrieval: SciFact dataset\cite{scifact}; 4).Clustering: BioRxiv clustering dataset\cite{mteb_cluster}.

\textbf{Evaluation metrics}:
To assess whether enhancing mutual information through our representation-focused approach improves general language understanding capabilities, we utilize multiple metrics: 1).Classification: accuracy, precision, recall, and F1 score; 2).Semantic Textual Similarity: Pearson and Spearman correlation; 3).Retrieval: Precision@5, Recall@5, F1@5, and MRR@5; 4).Clustering: Normalized Mutual Information (NMI) and Adjusted Rand Index (ARI).
% For each task, we use identical evaluation procedures and metrics for both the original and optimized models to ensure fair comparison. Improvements in performance across these diverse tasks would provide strong evidence that enhancing mutual information through our representation-focused approach leads to better general language understanding capabilities.

\begin{table*}[t]
    \caption{Token recoverability performance comparison across seven different LLMs on four different text datasets. Performance is measured using multiple metrics. The embedding decoder was trained on the Depedia dataset and then used to recover tokens from embeddings produced by each model, demonstrating differences in information preservation level across model architectures. Best performance for each metric and length category is highlighted in \textcolor{red}{red} (highest) and \textcolor{blue}{blue} (second highest).}
    \begin{center}
    \begin{sc}
    \setlength{\heavyrulewidth}{1.5pt}
    \setlength{\lightrulewidth}{1.2pt}
    \begin{small}
     \resizebox{\textwidth}{!}{
    \begin{tabular}{c|c|cc|ccc|cc}
    \toprule
       \textbf{Dataset}& \textbf{Model}& cos similarity & token f1 & bleu-1 & bleu-2 & bleu-4 & rouge-1 & rouge-L \\ 
       \midrule
         AG\_News&RoBERTa-large & \textcolor{red}{0.9544} & \textcolor{blue}{0.8765} & \textcolor{red}{0.8216} & \textcolor{red}{0.7621} & \textcolor{red}{0.6579} & \textcolor{blue}{0.8937} & \textcolor{blue}{0.8931} \\ 
         &DistilBERT & 0.8898 & \textcolor{red}{0.8877} & 0.5441 & 0.4697 & 0.3551 & \textcolor{red}{0.9520} & \textcolor{red}{0.9520} \\ 
         \hline
        &OPT & \textcolor{blue}{0.9484} & 0.8176 &\textcolor{blue}{0.7613} & \textcolor{blue}{0.6811} & \textcolor{blue}{0.5481} & 0.8525 & 0.8513 \\ 
        &Mistral & 0.9372 & 0.7821 & 0.7300 & 0.6433 & 0.5032 & 0.7975 & 0.7961 \\ 
        &Falcon & 0.9163 & 0.7059 & 0.6573 & 0.5528 & 0.3889 & 0.7325 & 0.7316 \\ 
        &GPT-NEO & 0.9368 & 0.7679 & 0.7041 & 0.6083 & 0.4535 & 0.7910 & 0.7899 
        \\ 
        & LLaMA2 & 0.9309 & 0.7685& 0.7142&0.6206&0.4712&0.7946&0.7944\\
        \midrule
        IMDB&RoBERTa-large & \textcolor{red}{0.9815} & \textcolor{blue}{0.8563} & \textcolor{red}{0.7933} & \textcolor{red}{0.7414} & \textcolor{red}{0.6485} & \textcolor{blue}{0.8936} & \textcolor{blue}{0.8916} \\ 
        &DistillBERT & 0.9565 & \textcolor{red}{0.8751} & 0.6217 & 0.5557 & 0.4435 & \textcolor{red}{0.9405} & \textcolor{red}{0.9402} \\ 
        \hline
        &OPT & \textcolor{blue}{0.9808} & 0.8071 & \textcolor{blue}{0.7432} & 0.6706 & 0.5488 & 0.8593 & 0.8559 \\ 
        &Mistral & 0.9780 & 0.7941 & 0.7397 & \textcolor{blue}{0.6744} & \textcolor{blue}{0.5635} & 0.8337 & 0.8303 \\  
        &Falcon & 0.9720 & 0.7101 & 0.6814 & 0.5896 & 0.4427 & 0.7824 & 0.7779 \\ 
        &GPT-NEO & 0.9748 & 0.7641 & 0.6934 & 0.6086 & 0.4716 & 0.8087 & 0.8020 \\ 
        &LLaMA2 & 0.9772& 0.7832& 0.7235&0.6513&0.5299&0.8253&0.8221\\
        \midrule
        
        QNLI & RoBERTa-large& \textcolor{blue}{0.9622} & \textcolor{blue}{0.8925} & \textcolor{red}{0.8383} & \textcolor{red}{0.7847} & \textcolor{red}{0.6840} & \textcolor{blue}{0.9264} & 0.9258 \\
        &DistillBERT & 0.9135 &\textcolor{red}{0.8931} & 0.6393 & 0.5683 & 0.4481 & \textcolor{red}{0.9666} & \textcolor{red}{0.9666} \\ 
        \hline
        &OPT & \textcolor{red}{0.9654} & 0.8779 & \textcolor{blue}{0.8267} & \textcolor{blue}{0.7669} & \textcolor{blue}{0.6542} & 0.9261 & \textcolor{blue}{0.9261} \\ 
        &Mistral & 0.9570 & 0.8595 & 0.8209 & 0.7539 & 0.6329 & 0.8982 & 0.8980 \\ 
        &Falcon & 0.9360 & 0.7990 & 0.7610 & 0.6777 & 0.5348 & 0.8424 & 0.8420 \\ 
        &GPT-NEO & 0.9556 & 0.8440 & 0.7938 & 0.7195 & 0.5904 & 0.8857 & 0.8846 \\ 

        & LLaMA2 & 0.9542& 0.8508& 0.8144 &0.7467&0.6274& 0.8844 &0.8841\\
        \midrule
        
        Chatgpt\_response &RoBERTa-large & \textcolor{red}{0.9711} & \textcolor{blue}{0.8514} & \textcolor{red}{0.8040} & \textcolor{red}{0.7393} & \textcolor{red}{0.6283} & \textcolor{blue}{0.9047} & \textcolor{blue}{0.9033} \\
        &DistillBERT & 0.9516 & \textcolor{red}{0.8902} & 0.6788 & 0.6083 & 0.4886 & \textcolor{red}{0.9539} & \textcolor{red}{0.9538} \\ 
        \hline
        &OPT & \textcolor{blue}{0.9648} & 0.7899 & \textcolor{blue}{0.7301} & \textcolor{blue}{0.6391} & \textcolor{blue}{0.4940} & 0.8549 & 0.8523 \\ 
        &Mistral & 0.9625 & 0.7527 & 0.7061 & 0.6154 & 0.4715 & 0.8102 & 0.8065 \\
        &Falcon & 0.9496 & 0.6716 & 0.6550 & 0.5411 & 0.3739 & 0.7604 & 0.7563 \\ 
        &GPT-NEO & 0.9594 & 0.7413 & 0.6815 & 0.5765 & 0.4155 & 0.8016 & 0.7947 \\ 
        &LLaMA2 & 0.9620& 0.7711 &0.7118& 0.6188 &0.4708 &0.8211 &0.8176\\
        \bottomrule
           
    \end{tabular}}
    \end{small}
    \end{sc}
    \end{center}
    \label{performance_different_domain_yelp}
\end{table*}

\section{Experiments}
In this section, we present a comprehensive evaluation of our information-theoretic framework through a series of experiments. We first examine token-level recoverability across different aspects: comparing various LLM architectures on texts from diverse domains (Section \ref{diverse_domain}), information flow patterns across layers (Section \ref{layer_wise}), investigating how text length affects mutual information (Section \ref{different_length}), and validate our theoretical framework's practical application in Section \ref{understanding_enhance}.
\subsection{Recoverability for texts from different domains}
\label{diverse_domain}
We evaluated token-level recoverability performance across seven different LLMs on texts from four different domains. As shown in Table~\ref{performance_different_domain_yelp}, encoder-only LLMs consistently demonstrate superior performance compared to decoder-only architectures, despite the significant parameter advantage of advanced decoder-only models. Across nearly all datasets and metrics, encoder-only models (RoBERTa-large and DistilBERT) achieve top-tier performance, aligned with the inherent advantages of their bidirectional attention mechanisms, which allow these models to simultaneously consider both left and right contexts when encoding token representation. DistilBERT demonstrates exceptional token-level recoverability with the highest Token F1 scores across all domains (ranging from 0.8751 to 0.8931). On BLEU metrics, RoBERTa-large maintains a clear advantage and DistilBERT shows remarkable strength on ROUGE metrics, achieving the highest scores across all domains (up to 0.9666 on QNLI). Among decoder-only models, OPT consistently performs best, occasionally reaching competitive levels with encoder-only models. Mistral typically ranks as the second-best decoder-only model, while Falcon generally shows the lowest performance across all architectures tested. These findings strongly indicate that encoder-only LLMs preserve richer information in their contextual embeddings compared to decoder-only counterparts. According to our theoretical analysis, this enhanced information preservation directly correlates with superior sentence understanding capabilities, aligning well with previous empirical observations~\cite{exp_encoder_decoder_2}.

\begin{figure}
  \centering
  \begin{subfigure}{0.48\textwidth}
    \centering
    \includegraphics[width=\linewidth,scale=0.8]{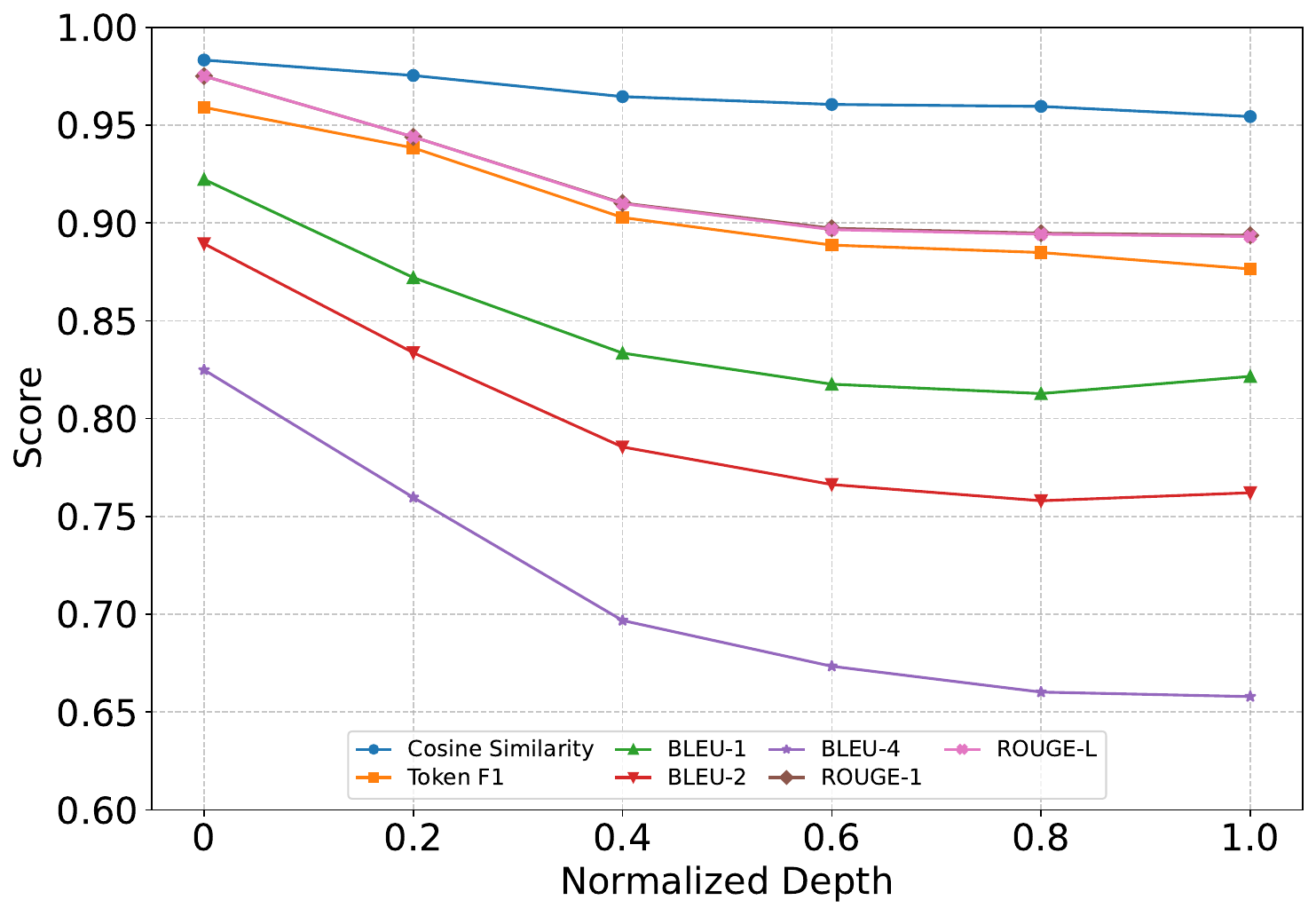}
    \caption{RoBERTa-large.}
    \label{roberta_trend}
  \end{subfigure}
  \hfill
  \begin{subfigure}{0.48\textwidth}
    \centering
    \includegraphics[width=\linewidth,scale=0.8]{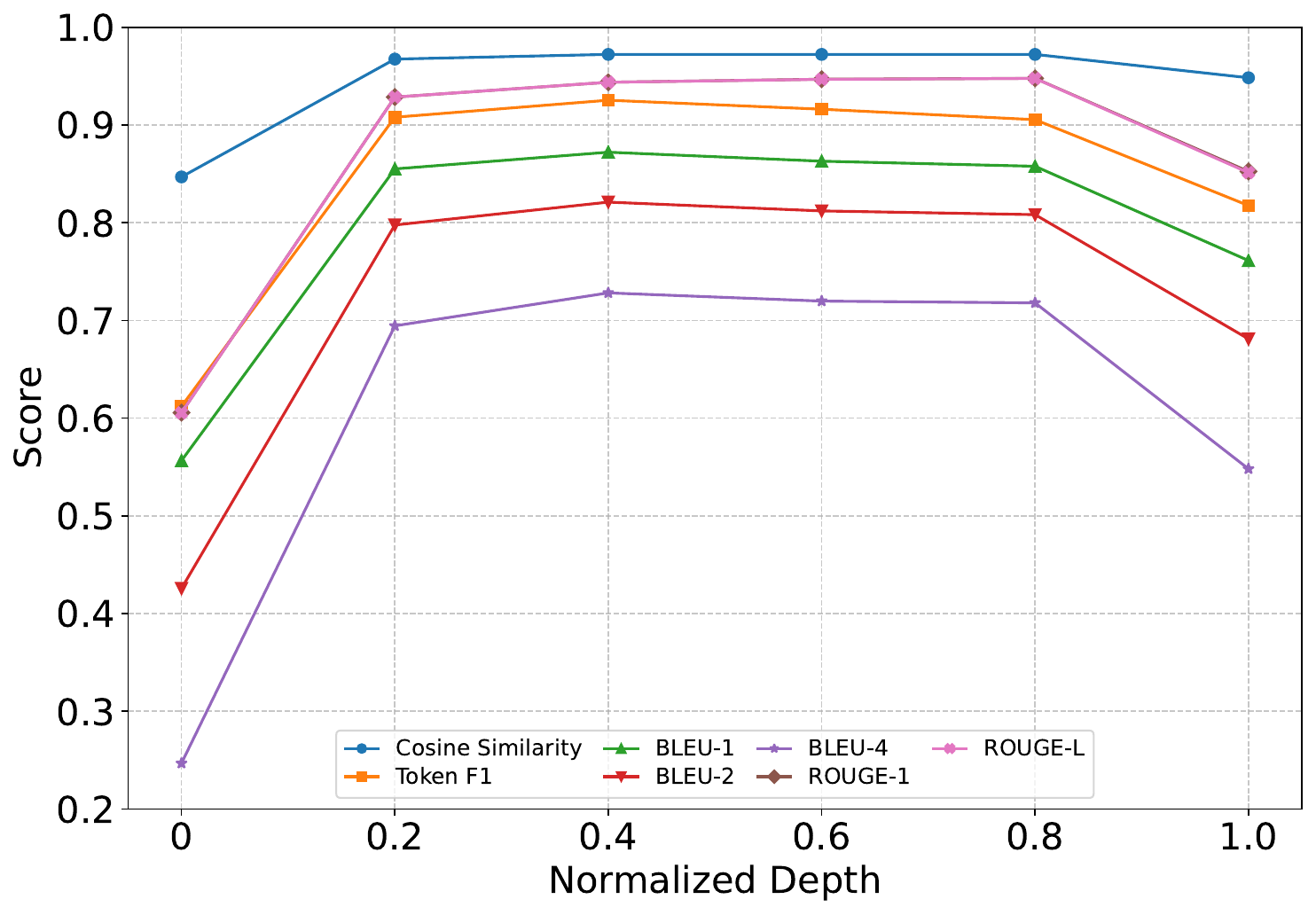}
    \caption{OPT-2.7b.}
    \label{opt_trend}
  \end{subfigure}
  \caption{Layer recoverability performance across model depths for RoBERTa-large (encoder-only) and OPT-2.7b (decoder-only) architectures. The plots show scores for various evaluation metrics at different normalized layer depths.}
  \label{fig:model_comparison}
  \vskip -0.1in
\end{figure}

\subsection{Recoverability across different layers}
\label{layer_wise}
To investigate information flow across model architectures, we analyzed token recoverability patterns in RoBERTa-large (encoder-only) and OPT (decoder-only). As shown in Fig.~\ref{fig:model_comparison}, these architectures exhibit fundamentally different patterns. RoBERTa-large demonstrates superior base representations and shows a gradual decline in recoverability from first to last layer, suggesting a progressive transformation toward contextualized representations while preserving semantic content. In contrast, OPT displays a distinctive "inverted U-shape" pattern—metrics increase sharply from the base layer, peak at intermediate layers, then decline toward the output layer. This reveals a "forgetting" process where token-specific information is first enhanced, then selectively discarded as the model prioritizes next-token prediction, aligned with previous study\cite{median_layer}. These architectural differences align with their distinct purposes: encoder-only models optimize for understanding entire sequences, while decoder-only models balance understanding with generation capabilities.

\begin{table*}[t]
    \caption{Token recoverability performance across different LLM architectures on on four text length categories: short (50-100 tokens), medium (100-300 tokens), long (300-800 tokens), and very long (800-1500 tokens). Performance is measured using multiple metrics. Best performance for each metric and length category is highlighted in \textcolor{red}{red} (highest) and \textcolor{blue}{blue} (second highest).}
    \begin{center}
    \begin{sc}
    \setlength{\heavyrulewidth}{1.5pt}
    \setlength{\lightrulewidth}{1.2pt}
    \begin{small}
     \resizebox{\textwidth}{!}{
    \begin{tabular}{c|c|cc|ccc|cc}
    \toprule
         && cos similarity & token f1 & bleu-1 & bleu-2 & bleu-4 & rouge-1 & rouge-L \\
        \midrule        
        amazon<100 &  RoBERTa-large & \textcolor{red}{0.9795} & \textcolor{red}{0.8995} & \textcolor{red}{0.8588} & \textcolor{red}{0.8113} & \textcolor{red}{0.7255} & \textcolor{blue}{0.9274} & \textcolor{blue}{0.9271} \\
        &DistilBERT & 0.9602 & \textcolor{blue}{0.8951} & 0.7033 & 0.6340 & 0.5124 & \textcolor{red}{0.9656} & \textcolor{red}{0.9656} \\
        \hline
        &OPT & \textcolor{blue}{0.9750} & 0.8365 & 0.7928 & 0.7188 & 0.5921 & 0.8719 & 0.8710 \\ 
        &Mistral & 0.9746 & 0.8362 & \textcolor{blue}{0.8024} & \textcolor{blue}{0.7356} & \textcolor{blue}{0.6207} & 0.8686 & 0.8681 \\  
        &Falcon & 0.9636 & 0.7586 & 0.7338 & 0.6378 & 0.4828 & 0.8098 & 0.8085 \\
        &GPT-NEO & 0.9689 & 0.7866 & 0.7371 & 0.6462 & 0.4988 & 0.8189 & 0.8171 \\
        &LLaMA2 & 0.9739 & 0.8248& 0.7858 & 0.7119 & 0.5863 &0.8545 &0.8537\\
        \midrule
        
        100<yelp<300 & RoBERTa-large & \textcolor{red}{0.9810} & \textcolor{blue}{0.8639} & \textcolor{red}{0.8403} & \textcolor{red}{0.7832} & \textcolor{red}{0.6801} & \textcolor{blue}{0.8926} & \textcolor{blue}{0.8923} \\ 
        &DistilBERT & 0.9685 & \textcolor{red}{0.8783} & 0.6884 & 0.6185 & 0.4955 & \textcolor{red}{0.9648} & \textcolor{red}{0.9648} \\ \hline
        &OPT & 0.9753 & 0.7947 & 0.7617 & 0.6747 & 0.5310 & 0.8319 & 0.8291 \\ 
        &Mistral & \textcolor{blue}{0.9792} & 0.7950 & \textcolor{blue}{0.7750} & \textcolor{blue}{0.6978} & \textcolor{blue}{0.5667} & 0.8413 & 0.8382 \\ 
        &Falcon & 0.9673 & 0.6824 & 0.6855 & 0.5728 & 0.3998 & 0.7581 & 0.7529 \\ 
        &GPT-NEO & 0.9724 & 0.7349 & 0.6842 & 0.5776 & 0.4153 & 0.7677 & 0.7610 \\ 
        &LLaMA2 & 0.9756 & 0.7645 & 0.7413 & 0.6525 &0.5063 &0.8083 & 0.8052\\
        \midrule
        
        300<yelp<800 & RoBERTa-large & \textcolor{red}{0.9853} & \textcolor{blue}{0.8015} & \textcolor{red}{0.7455} & \textcolor{red}{0.6899} & \textcolor{red}{0.5940} & \textcolor{blue}{0.8407} & \textcolor{blue}{0.8352} \\
        &DistilBERT & 0.9693 & \textcolor{red}{0.8305} & 0.6386 & 0.5690 & 0.4543 & \textcolor{red}{0.9001} & \textcolor{red}{0.8990} \\ \hline
        &OPT & 0.9797 & 0.7455 & \textcolor{blue}{0.6671} & 0.5814 & 0.4468 & 0.7762 & 0.7622 \\ 
        &Mistral & \textcolor{blue}{0.9831} & 0.7298 & 0.6551 & \textcolor{blue}{0.5820} & \textcolor{blue}{0.4630} & 0.7698 & 0.7593 \\ 
        &Falcon & 0.9751 & 0.6292 & 0.6015 & 0.4947 & 0.3395 & 0.7077 & 0.6898 \\ 
        &GPT-NEO & 0.9779 & 0.6824 & 0.6087 & 0.5019 & 0.3493 & 0.7198 & 0.6950 \\ 

         &LLaMA2 & 0.9798 & 0.6911 & 0.6184 & 0.5356 & 0.4062 & 0.7332 & 0.7179\\
        \midrule
        800<yelp<1500 & RoBERTa-large & \textcolor{red}{0.9860} & \textcolor{blue}{0.6231} & \textcolor{blue}{0.2719} & \textcolor{blue}{0.2511} & \textcolor{red}{0.2177} & \textcolor{blue}{0.5752} & \textcolor{blue}{0.5619} \\ 
        &DistilBERT & 0.9684 & \textcolor{red}{0.6372} & \textcolor{red}{0.2875} & \textcolor{red}{0.2511} & \textcolor{blue}{0.1981} & \textcolor{red}{0.5973} & \textcolor{red}{0.5937} \\ \hline
        &OPT & 0.9806 & 0.5903 & 0.2478 & 0.2126 & 0.1622 & 0.5459 & 0.5115 \\ 
        &Mistral & \textcolor{blue}{0.9839} & 0.5741 & 0.2237 & 0.1977 & 0.1583 & 0.5223 & 0.5012 \\ 
        &Falcon & 0.9770 & 0.5068 & 0.2242 & 0.1821 & 0.1252 & 0.5011 & 0.4627 \\ 
        &GPT-NEO & 0.9794 & 0.5490 & 0.2498 & 0.2006 & 0.1383 & 0.5269 & 0.4706 \\
        &LLaMA2 & 0.9809 & 0.5449 & 0.2048 & 0.1760 &0.1345 & 0.4974 & 0.4695\\
        \bottomrule
    \end{tabular}}
    \end{small}
    \end{sc}
    \end{center}
    \label{text_different_length}
\end{table*} 
\subsection{Recoverability on texts with different lengths}
\label{different_length}
To investigate how text length affects mutual information across different models, we conducted experiments with texts of varying lengths. 
As shown in Table~\ref{text_different_length}, for shorter texts (50-100 tokens), both encoder-only and decoder-only models demonstrate strong performance. As the text length increases to the medium range (100-300 tokens), we observe a modest but consistent decline in performance across all models and metrics. The performance gap between encoder-only and decoder-only architectures becomes more pronounced, with RoBERTa-large maintaining significantly higher BLEU-4 scores compared to the best decoder-only model, Mistral. Performance degradation accelerates for long texts (300-800 tokens), particularly for decoder-only models. While RoBERTa-large maintains relatively strong performance, decoder-only models show a steeper decline, with even the best-performing Mistral dropping to 0.4630 on BLEU-4. For very long texts (800-1500 tokens), all models experience a dramatic performance drop, particularly on sequence-level metrics like BLEU and ROUGE. Notably, encoder-only models still maintain their advantage, with both RoBERTa-large and DistilBERT outperforming decoder-only LLMs. 

% As the length of the text increases, we observe a clear pattern of decreasing recovery, regardless of model architecture. This degradation pattern suggests the mutual information between individual tokens and the overall sentence embedding diminishes as input text length increases, making accurate token recovery more challenging. 
% \subsection{Recoverability across models in a series with different sizes}

\begin{table}
    \caption{Performance comparison between baseline LLMs and their mutual information-enhanced counterparts on semantic textual similarity (STS) and clustering tasks. The table presents results for three models before and after fine-tuning with our proposed token recoverability optimization approach. The results demonstrate consistent improvements across all models and metrics after mutual information enhancement.}
    \begin{center}
    \begin{sc}
    \setlength{\heavyrulewidth}{1.5pt}
    \setlength{\lightrulewidth}{1.2pt}
    \begin{small}
    \resizebox{\linewidth}{!}{
    \begin{tabular}{c|cc|cc|cc|cc}
    \toprule
        \multirow{3}{*}{\textbf{Model}} & \multicolumn{4}{c|}{\textbf{Original}} & \multicolumn{4}{c}{\textbf{Enhanced}} \\ 
        \cline{2-9}
        & \multicolumn{2}{c|}{STS} & \multicolumn{2}{c|}{Clustering} & \multicolumn{2}{c|}{STS} & \multicolumn{2}{c}{Clustering} \\ 
        \cline{2-9}
        & pearson & spearman & NMI & ARI & pearson & spearman & NMI & ARI \\ 
        \hline
        opt-2.7b & 0.8778 & 0.8792 & 0.1566 & 0.0758 & 0.8784 $\uparrow$ & 0.8798 $\uparrow$ & 0.1589 $\uparrow$ & 0.0770 $\uparrow$ \\ 
        llama2-7b & 0.8571 & 0.8614 & 0.1064 & 0.0458 & 0.8590 $\uparrow$ & 0.8626 $\uparrow$ & 0.1560 $\uparrow$ & 0.0975 $\uparrow$ \\ 
        mistral-7b & 0.8910 & 0.8919 & 0.0341 & 0.0086 & 0.9090 $\uparrow$ & 0.9090 $\uparrow$ & 0.1435 $\uparrow$ & 0.0583 $\uparrow$ \\ 
        \bottomrule
    \end{tabular}
    }
    \end{small}
    \end{sc}
    \end{center}
    \label{tab:sts_clustering_performance}
\end{table}

\begin{figure}
  \centering\includegraphics[width=\linewidth]{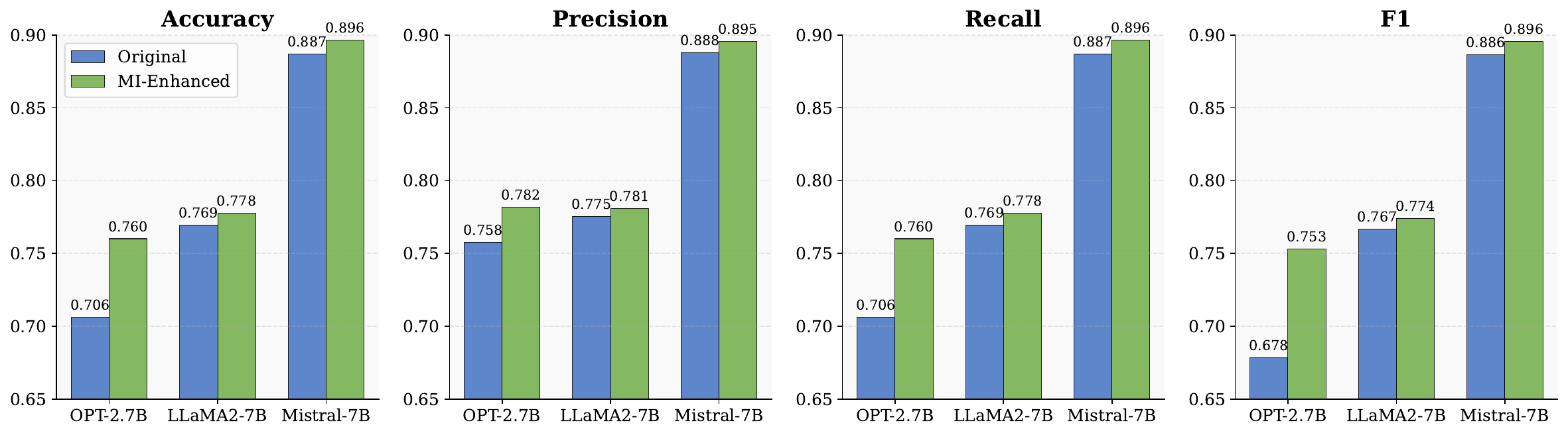}
  \caption{Classification performance comparison between original models and their mutual information-enhanced versions on the AG-News dataset. The bar charts show improvements across four key metrics (Accuracy, Precision, Recall, and F1 score) for three different LLMs (OPT-2.7B, LLaMA2-7B, and Mistral-7B).}
  \label{classification}
\end{figure}

\begin{figure}
  \centering\includegraphics[width=\linewidth]{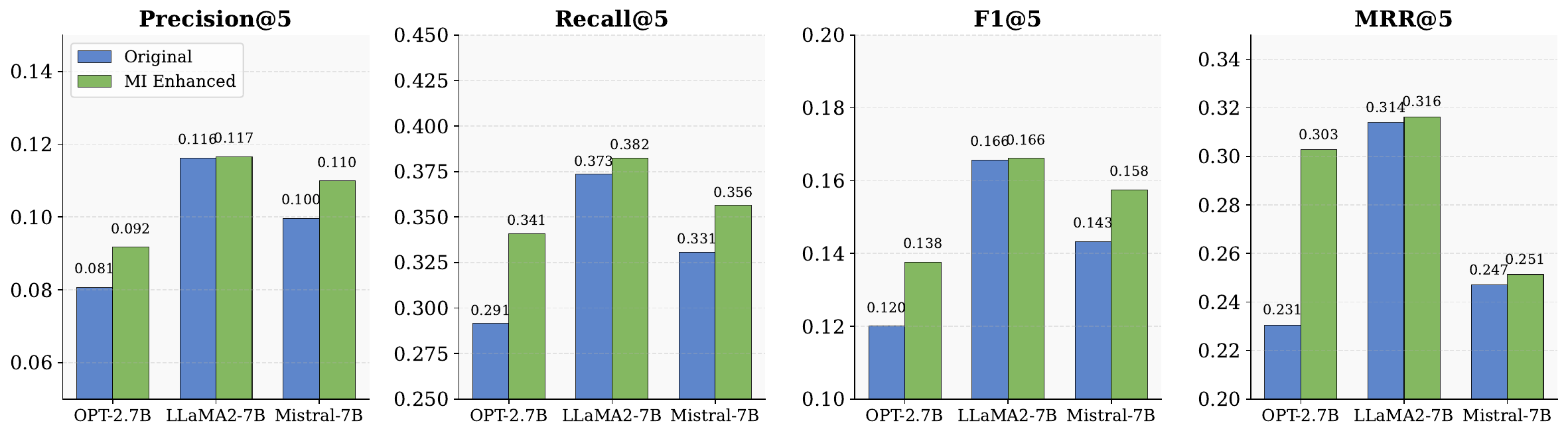}
  \caption{Retrieval performance comparison between original models and their mutual information-enhanced versions. The bar charts display four retrieval metrics (Precision@5, Recall@5, F1@5, and MRR@5) for three different LLMs (OPT-2.7B, LLaMA2-7B, and Mistral-7B).}
  \label{retrieval}
\end{figure}

\subsection{Understanding ability enhanced results}
\label{understanding_enhance}
To empirically validate our theoretical framework empirically, we conducted experiments to enhance LLM understanding ability through fine-tuning. We evaluated three LLMs: OPT-2.7B, LLaMA2-7B, and Mistral-7B across various information retrieval, semantic textual similarity, and clustering tasks.\\
\textbf{Retrieval}:
As illustrated in Figure~\ref{retrieval}, all models exhibited substantial improvements in information retrieval capabilities after MI enhancement training. For Precision@5, OPT-2.7B showed the largest relative gain from 0.081 to 0.092. Similar patterns were observed for Recall@5, where LLaMA2-7B maintained the highest absolute performance, and OPT-2.7B exhibited the most substantial relative improvement from 0.291 to 0.341. The F1@5 metric, which balances precision and recall, further confirms this trend, with LLaMA2-7B achieving the highest score (0.166) and OPT-2.7B showing the most significant relative improvement (from 0.120 to 0.138). For MRR@5, OPT-2.7B demonstrated an improvement from 0.231 to 0.303, while LLaMA2-7B maintained the highest absolute performance (0.316).\\
\textbf{Classification}:
As shown in Figure~\ref{classification}, all models exhibited substantial improvements in classification performance after MI enhancement training. While OPT-2.7B showed the largest relative improvement, with its accuracy increasing from 0.706 to 0.760. Mistral-7B maintained the highest absolute performance with an accuracy of 0.896 after enhancement.
The enhanced models consistently outperformed their original counterparts across precision, recall, and F1 metrics. In particular, OPT-2.7B exhibited the most substantial improvement in the F1 score (from 0.678 to 0.753). \\
\textbf{Semantic textual similarity and clustering}:
Table~\ref{tab:sts_clustering_performance} presents performance improvements on semantic textual similarity (STS) and clustering tasks. For STS, all models demonstrate improved correlation after MI optimization. Mistral-7B showed the most substantial improvement in both Pearson and Spearman correlation coefficients.
The clustering results reveal even more dramatic improvements. LLaMA2-7B exhibited the most substantial gains in cluster quality, with NMI increasing from 0.1064 to 0.1560 and ARI more than doubling from 0.0458 to 0.0975. Remarkably, Mistral-7B, which initially showed poor clustering performance, demonstrated a dramatic improvement after enhancement with its ARI increasing by nearly 7×.

\section{Conclusion}

Our work established an information-theoretic framework for evaluating LLM understanding capabilities using mutual information. Through extensive experiments, we demonstrated that encoder-only LLMs keep a high mutual information than decoder-only LLMs, which exhibit a distinctive "forgetting" process in the latter. By fine-tuning models to enhance token-level mutual information, we achieved consistent improvements across various downstream tasks, confirming that mutual information serves as a fundamental building block for language understanding. Our framework offers an architecture-agnostic evaluation paradigm that provides deeper insight into model behavior than traditional task-specific evaluation. Our findings highlight how architectural choices influence information processing in LLMs and suggest promising directions for developing more effective language models.

%%%%%%%%%%%%%%%%%%%%%%%%%%%%%%%%%%%%%%%%%%%%%%%%%%%%%%%%%%%%
\newpage
\appendix

\section{Proof for relationship between sentence-level and token-level mutual information}
\label{MI_sentence_token}
\begin{proof}
We start with the definition of mutual information between the sentence $S = (t_1, t_2, \ldots, t_n)$ and its embedding $E$:
\begin{equation}
I(S; E) = H(S) - H(S|E)
\end{equation}
For individual tokens, the average mutual information is:
\begin{equation}
\frac{1}{n}\sum_{i=1}^{n} I(t_i; E) = \frac{1}{n}\sum_{i=1}^{n} [H(t_i) - H(t_i|E)]
\end{equation}
Multiplying both sides by $n$:
\begin{equation}
n \cdot \frac{1}{n}\sum_{i=1}^{n} I(t_i; E) = \sum_{i=1}^{n} H(t_i) - \sum_{i=1}^{n} H(t_i|E)
\end{equation}
Now, by the chain rule of entropy:
\begin{equation}
H(S|E) = H(t_1, t_2, \ldots, t_n|E) = \sum_{i=1}^{n} H(t_i|E, t_1, \ldots, t_{i-1})
\end{equation}
Due to the data processing inequality and the fact that conditioning reduces entropy:
\begin{equation}
H(t_i|E) \geq H(t_i|E, t_1, \ldots, t_{i-1})
\end{equation}
Therefore:
\begin{equation}
\sum_{i=1}^{n} H(t_i|E) \geq \sum_{i=1}^{n} H(t_i|E, t_1, \ldots, t_{i-1}) = H(S|E)
\end{equation}
Substituting this inequality:
\begin{equation}
n \cdot \frac{1}{n}\sum_{i=1}^{n} I(t_i; E) = \sum_{i=1}^{n} H(t_i) - \sum_{i=1}^{n} H(t_i|E) \leq \sum_{i=1}^{n} H(t_i) - H(S|E)
\end{equation}
Rearranging:
\begin{equation}
H(S|E) \leq \sum_{i=1}^{n} H(t_i) - n \cdot \frac{1}{n}\sum_{i=1}^{n} I(t_i; E)
\end{equation}
Returning to our original mutual information definition:
\begin{align}
I(S; E) &= H(S) - H(S|E) \
&\geq H(S) - \left[\sum_{i=1}^{n} H(t_i) - n \cdot \frac{1}{n}\sum_{i=1}^{n} I(t_i; E)\right] \ 
&= H(S) - \sum_{i=1}^{n} H(t_i) + n \cdot \frac{1}{n}\sum_{i=1}^{n} I(t_i; E)
\end{align}
Since $H(S) = H(t_1, t_2, \ldots, t_n)$, we have:
\begin{equation}
I(S; E) \geq n \cdot \frac{1}{n}\sum_{i=1}^{n} I(t_i; E) + H(t_1, t_2, \ldots, t_n) - \sum_{i=1}^{n} H(t_i)
\end{equation}

let $\Delta_=H(t_1, t_2, \ldots, t_n) - \sum_{i=1}^{n} H(t_i)$, we get:
\begin{equation}
I(S; E) \geq n \cdot \frac{1}{n}\sum_{i=1}^{n} I(t_i; E) + \Delta
\end{equation}
\end{proof}

\section{Proof for relationship between recoverability and Mutual information}
\label{MI_rec}
\begin{proof}
We begin by considering the recovery of individual tokens. Let $t_i$ be the $i$-th token in sentence $S$, and let $P_{e,i} = 1 - \mathbb{P}[f(E)_i = t_i]$ be the probability of error in recovering token $t_i$ from embedding $E$ using the optimal decoder $f$.

By Fano's inequality applied to each token individually, we have:
\begin{equation}
H(t_i|E) \leq H_b(P_{e,i}) + P_{e,i}\log(|\mathcal{V}| - 1)
\end{equation}
where $H(t_i|E)$ is the conditional entropy of token $t_i$ given embedding $E$.

Since the binary entropy function is symmetric, i.e., $H_b(P_{e,i}) = H_b(1-P_{e,i}) = H_b(\mathbb{P}[f(E)_i = t_i])$, we can rewrite this as:
\begin{equation}
H(t_i|E) \leq H_b(\mathbb{P}[f(E)_i = t_i]) + (1-\mathbb{P}[f(E)_i = t_i])\log(|\mathcal{V}| - 1)
\end{equation}

By the definition of mutual information for each token:
\begin{equation}
I(t_i;E) = H(t_i) - H(t_i|E)
\end{equation}

Since we assume tokens are uniformly distributed over $\mathcal{V}$, we have $H(t_i) = \log |\mathcal{V}|$. Then:
\begin{align}
I(t_i;E) &= \log |\mathcal{V}| - H(t_i|E) \notag  \\
&\geq \log |\mathcal{V}| - [H_b(\mathbb{P}[f(E)_i = t_i]) + (1-\mathbb{P}[f(E)_i = t_i])\log(|\mathcal{V}| - 1)] \notag  \\
&= \log |\mathcal{V}| - H_b(\mathbb{P}[f(E)_i = t_i]) - (1-\mathbb{P}[f(E)_i = t_i])\log(|\mathcal{V}| - 1)
\end{align}

Rearranging terms:
\begin{align}
I(t_i;E) &\geq \log |\mathcal{V}| - H_b(\mathbb{P}[f(E)_i = t_i]) - \log(|\mathcal{V}| - 1) + \mathbb{P}[f(E)_i = t_i]\log(|\mathcal{V}| - 1) \notag  \\
&= \mathbb{P}[f(E)_i = t_i]\log(|\mathcal{V}| - 1) - H_b(\mathbb{P}[f(E)_i = t_i]) + \log |\mathcal{V}| - \log(|\mathcal{V}| - 1)
\end{align}

For large $|\mathcal{V}|$ (where $\log(|\mathcal{V}|-1) \approx \log|\mathcal{V}|$), the term $\log |\mathcal{V}| - \log(|\mathcal{V}| - 1)$ approaches zero, so this inequality simplifies to:
\begin{equation}
I(t_i;E) \geq \mathbb{P}[f(E)_i = t_i]\log(|\mathcal{V}| - 1) - H_b(\mathbb{P}[f(E)_i = t_i])
\end{equation}

Taking the average across all tokens in the sentence:
\begin{align}
\frac{1}{n}\sum_{i=1}^{n} I(t_i;E) &\geq \frac{1}{n}\sum_{i=1}^{n} \left[ \mathbb{P}[f(E)_i = t_i]\log(|\mathcal{V}| - 1) - H_b(\mathbb{P}[f(E)_i = t_i]) \right]
\end{align}

By Jensen's inequality and the concavity of the binary entropy function, we have:
\begin{equation}
\frac{1}{n}\sum_{i=1}^{n}H_b(\mathbb{P}[f(E)_i = t_i]) \leq H_b\left(\frac{1}{n}\sum_{i=1}^{n}\mathbb{P}[f(E)_i = t_i]\right) = H_b(P_{\mathrm{rec}}(S, E))
\end{equation}

Therefore:
\begin{align}
\frac{1}{n}\sum_{i=1}^{n} I(t_i;E) &\geq \frac{1}{n}\sum_{i=1}^{n} \mathbb{P}[f(E)_i = t_i]\log(|\mathcal{V}| - 1) - \frac{1}{n}\sum_{i=1}^{n} H_b(\mathbb{P}[f(E)_i = t_i]) \notag  \\
&\geq \frac{1}{n}\sum_{i=1}^{n} \mathbb{P}[f(E)_i = t_i]\log(|\mathcal{V}| - 1) - H_b(P_{\mathrm{rec}}(S, E)) \notag  \\
&= P_{\mathrm{rec}}(S, E)\log(|\mathcal{V}| - 1) - H_b(P_{\mathrm{rec}}(S, E))
\end{align}

% This establishes our lower bound on the average token-level mutual information in terms of token-level recoverability.
\end{proof}

\bibliographystyle{plain}
\bibliography{neurips_2025}

\end{document}